# Automatic Alignment of English-Chinese Bilingual Texts of CNS News

Donghua Xu — xudonghu@iscs.nus.sg
Chew Lim Tan — tancl@iscs.nus.sg
Department of Information System and Computer Science,
National University of Singapore, Singapore 119260

## Abstract

In this paper we address a method to align English-Chinese bilingual news reports from China News Service, combining both lexical and statistical approaches. Because of the sentential structure differences between English and Chinese, matching at the sentence level as in many other works may result in frequent matching of several sentences en masse. In view of this, the current work also attempts to create shorter alignment pairs by permitting finer matching between clauses from both texts if possible. The current method is based on statistical correlation between sentence or clause length of both texts and at the same time uses obvious anchors such as numbers and place names appearing frequently in the news reports as lexical cues.

## 1. Introduction

Example-based Machine Translation proposed by Nagao[1][2] has received increasing attention from researchers in recent years. The example-based approach, however, needs a huge collection of translation examples. Automatic alignment of bilingual texts will thus prove to be very useful in facilitating the creation of a large example-base for the machine translation.

Several methods of automatic alignment[3]-[8] have already been reported in the literature. Among them, we find basically three different approaches, namely, the lexical approach, the statistical approach, and the combination of them. The lexical approach as in the works of Kay and Roscheisen[3][4] and of Hwang and Nagao[8], looks for relationship between lexical contents of the bilingual texts in order to find alignment pairs. The statistical approach as demonstrated in the methods by Brown et al.[5] and By Gale and Church[6][7], on the other hand, uses statistical correlation between sentence lengths of the bilingual texts as the basis of matching. And some recent works as presented by Wu[9] and by Tan[10] et al. combine both approaches to align texts.

In our present work, we would like to study the alignment between the English-Chinese bilingual news reports from China News Service (CNS). The next section will discuss some features of these texts.

## 2. Features of CNS News

With the recent rapid development of Internet and World Wide Web (WWW), we can now easily retrieve a large volume of linguistic resources that were not so easy to get in computer format before. The news reports released by the news agencies and other organizations, because of their large amount everyday and relatively formal expression, will prove to be very valuable for researchers on computational linguistics.

China News Service is one of the two major news agencies in China. It offers news from China in both Chinese and English on WWW. On their Web page, each file contains a single piece of news in one language, so the files are fairly small (usually 1K or 2K per file). Matching between corresponding files will not require too deep recursion of Dynamic Programming. This seems to be an advantage for our automatic alignment. However, the separated files make it a bit complicated to download. We will discuss this later.

Each news file consists of a number of paragraphs separated by clear delimiters. Each paragraph is composed of some complete sentences which end with “。 。 。 ” in Chinese and “.!?” in English. And each sentence contains a number of clauses divided by “， ， ， ” both in Chinese and in English. Of course in English texts “.” and “,” do not necessarily mean the end of a sentence and a clause respectively. The former might be a decimal point of a number or a mark of abbreviation, while the latter might be a delimiter at every three consecutive digits of a number. Fortunately, in the news reports we are studying, all the digits of a complete number are immediately adjacent, thus we can simply check whether the characters before and after “.”(“,”) are both digits to decide whether the “.”(“,”) is a part of a number. And the abbreviations with a “.” such as “U.S.” or “Mr.” can be picked out by a table look-up. For simplicity of discussion, we will call the end punctuation of sentence (i.e. “。 .!?”) “period”, and that of clause (i.e. “,;:”) “comma” from now on.

After extensive observations, we determine that there are only two kinds of match between paragraphs in the two languages, i.e. $1 \Leftrightarrow n$ and $n \Leftrightarrow 1$, the former means 1 English paragraph matches with $n$ successive Chinese paragraphs, and the latter vice-versa. It depends on the human translator. Since the corpora are translated from Chinese into English, sometimes the translator would break a Chinese paragraph which contains some collateral or irrelative sentences into several collateral English paragraphs, while sometimes combine several collateral Chinese paragraphs into one English paragraph. On the other hand, we found that some of the Chinese paragraphs that might be less important are not translated at all, so we define $0 \Leftrightarrow n$ as another kind of match, i.e. 0 English paragraph match with $n$ consecutive Chinese paragraphs.

A feature with the English and Chinese language pair is their differences in sentential structures. The differences are believed to be wider than the corpora pairs involved in the earlier work. The alignment methods by Kay and Roscheiscen[4], Brown et al.[5], and Gale and Church[6],[7] dealt with English, French and German, while Hwang and Nagao[8] handled Japanese and Korean texts. The languages studied in these works, as pointed out in Wu[9] and Tan[10], share cognates and similar grammatical structures. And in Wu[9], the corpora involved are Hong Kong Hansard, i.e. the record and the translation (between Cantonese and English) of the discourse on the Legislative Council, therefore the translation tends to be literal, despite the great difference between the two languages. They can get a high rate of $1 \Leftrightarrow 1$ match at nearly 90% when aligning on sentence level. In our case however, through a coarse estimation, we find that only about 50% of the sentences in CNS news can be matched as $1 \Leftrightarrow 1$. That means, if we just match the texts at sentence level, it would be very often to pair quite a number of sentences en masse between both texts. This is because of the differences between the language pair in the use of periods to end sentence.

Therefore, what we would like to propose is to allow a sentence in one text to be matched with part of a sentence of the other text if possible in order to create a finer correspondence pair. We observe that a period in either text indicates a self-contained entity and may be used to find a break point in the other text even if there is no period but a comma at that point. The sentence which ends with that period may then be matched with the sequence of clauses ended with that break point in the other text.

Matching clauses will be more difficult and error-prone than simple sentences matching. Fortunately, another useful feature of these news reports is the obvious anchors in the texts, e.g. digit numbers (“二二二” in Chinese and “23” or “twenty-three” in English), or places (“北北” in Chinese and “Beijing” in English), or specific dates (“十十十十” in Chinese and “October 1” in English), etc. There are plenty of these anchors in most of the news. The meaning of these anchors would be much less ambiguous than normal words, so we can easily find the match of these anchors in both texts and count the occurrence of these anchors, and make it an important element to decide how much the two strings of the texts are likely to match with each other.

Furthermore, we find that by measuring the degree of scattering of these anchors, we can even successfully achieve clause-to-clause match in some parts of the texts.

We divide the whole automatically aligning process into four steps:

1) Preprocess the English & Chinese corpora
2) Match the anchors in the two corpora
3) Align between paragraphs in English & Chinese corpora
4) Align between clauses of each pair of aligned paragraphs

These four steps are described one by one in the next section.

# 3. The Aligning Process

## 3.1 Preprocessing

In this step, the program reads in a pair of English and Chinese files and analyses them. The head and tail and some other noisy information of each news file are filtered out. A data structure is constructed for the paragraph-sentence-clause hierarchical organization of each corpus. And another important task of this step is to analyse the anchors in both corpora as discussed below.

First of all, the way to express a number is quite different in the two languages. For example, in Chinese the number “一一一一一一一一一一一” would be changed into “197.66 billion” in English translation. Another example is “七七七” (in Chinese “七” means 10%) which has the same meaning as “百百百百百百” and “百百百百百百百” and “四四四四”, and might appear as “75 per cent” or “three-fourths” in English texts. We have to make a conversion to transform a set of number-characters into a real number in the program considering all such cases in both texts, so that we can compare them later.

Secondly, since these news are from China, they are concerned with matters of China. Thus the places in China and countries in the world appear frequently in the news. We construct a table which contains the English and Chinese names of the countries in the world and provinces and big cities in China. We sort this table and use Binary Search to look it up when deciding whether a series of English or Chinese characters refer to a place. This table has now about 200 items, and thus with Binary Search, table look up can be done within 8 times comparison.

For each news file, we construct an Anchor Table to record all the anchors.

## 3.2 Matching Anchors

Having determined all the anchors in both corpora, we use each anchor in English Anchor Table to find a match in Chinese Anchor Table. Let $MA$ represent a pair of matched anchors. There are some interesting details in anchor matching worth mentioning here.

Firstly, we must give different weights to different types of anchors. An exact number such as "192.6" deserves a larger weight than a normal place name. This is because when translation is free, place names sometimes are used in different locations in the two languages, while exact numbers usually only appear in the corresponding sense group. So we define several types of anchors and specify the weight of each type of matched anchors $W(MA)$, for example:

$$W(MA) = \begin{cases} 100 & \text{if } MA \text{ is an exact number} \\ 10 & \text{if } MA \text{ is a place} \\ \cdots & \cdots \end{cases}$$

Secondly, when matching the anchors, we might find an anchor repeat several times in a corpus. For example, if there are 3 "北北" in the Chinese corpus and 2 "Beijing" in the English corpus, it is difficult for the program to directly decide which "北北" matches with which "Beijing". Therefore we adopt a special chain to record the repeated anchors, and allow all "北北" to match with all "Beijing" at a whole. And to prevent the repeated anchors from messing up the match process too much, we should give this kind of match a penalty by decreasing its weight. Thus we define the Factor of Repetition $F_R(MA)$ :

$$F_R(MA) = \begin{cases} 0.5 & \text{if } MA \text{ appears more than once} \\ 1 & \text{if } MA \text{ appears only once} \end{cases}$$

In addition, because the translation is free, sometimes an anchor in one language would be changed into an approximate form in another language. E.g. "一一一一一" in Chinese which means 196 might be changed into "nearly 200" in English. So if an anchor in English Anchor Table cannot find an exact match in the Chinese Anchor Table, we allow it to search again for an approximate match. Of course, an approximate match should raise another penalty. We define the Factor of Approximation $F_A(MA)$ as :

$$F_A(MA) = \begin{cases} 0.5 & \text{if } MA \text{ is an approximate match} \\ 1 & \text{if } MA \text{ is an exact match} \end{cases}$$

Therefore, the value of a pair of matched anchors is defined as:

$$V(MA) = W(MA) \cdot F_R(MA) \cdot F_A(MA)$$

For the sake of simplicity and clarity, we do not include some other processing details in this equation. For example, in Chinese "一" and 一一" which mean 1 and 2 respectively are often used as a part of a idiom or word, but not actual number. At the same time, "one" in English also is not always used as actual number. So we would not say a match between "一" and "one" is very reliable. Thus in this case the match should get a penalty too.

## 3.3 Aligning Paragraphs

After matching the anchors, we can proceed to align the two texts. In this step we treat paragraph as the basic unit of aligning, to find out which English paragraph(s) match with which Chinese paragraph(s).

Let $P_E$, $P_C$ be the total number of paragraphs respectively in the English corpus and the Chinese corpus. Let $match(u,v)$ represent the total cost of matching from $u$ th paragraph to the end of the English corpus with a part of the Chinese corpus from the $v$ th paragraph to the end, where $u = 1,2,\cdots,P_E$, $v = 1,2,\cdots,P_C$. In our program, we obtain $match(u,v)$ as follows:

$$match(u,v) = \min\{M(u,v), match(u,v+1)\} \quad (1)$$

Where $match(u,v+1)$ is the cost of ignoring the $v$ th Chinese paragraph. We have mentioned that sometimes some Chinese paragraphs are not translated into English at all. Thus besides $M(u,v)$ (i.e. the real cost of matching $u$ th English paragraph with $v$ th Chinese paragraph), we also compute $match(u,v+1)$ (i.e. the cost of matching $u$ th English paragraph with the next Chinese paragraph), and compare the two costs to choose the smaller one. If $match(u,v+1)$ is smaller, what we can determine is that the $v$ th Chinese paragraph matches with nothing, i.e. it is not translated. If so, however, the $u$ th English paragraph does not necessarily match with $(v+1)$ th paragraph, because when computing $match(u,v+1)$, through a similar process, we might find that $(v+1)$ th Chinese paragraph also matches with nothing. In this way we can find a match of $0 \Leftrightarrow n$, which means successive $n$ Chinese paragraphs match with nothing.

We assume that each English paragraph has its corresponding part in Chinese, so when there are still some English paragraphs not matched, we would not allow all the remaining Chinese paragraphs to match with nothing. Since the recursion works forward to the end of the corpora, when checking the terminate condition of the recursion at the end of each corpus, the program will follow different routines according to different conditions:

1)If it exceeds the end of the Chinese corpus but does not exceed that of the English corpus, i.e. there are some English paragraphs still not matched, this match must be a wrong match. So the program will return an error flag to the upper level of recursion indicating that this match is invalid.

2)Otherwise, i.e. if only it exceeds the end of the English corpus, no matter whether or not it exceeds that of the Chinese corpus, this match may be regarded as a valid match. So the program will return 0 as a lowest level cost to the upper level of recursion.

Returning to Eq.(1), $M(u,v)$ is the actual cost of matching the English text from the $u$ th paragraph to the end with the Chinese text from $v$ th paragraph to the end. Using Dynamic Programming, it is to achieve its minimum through a recursive process as follows:

$$M(u,v) = \min_{i=1}^{P_E-u} \min_{j=1}^{\beta(i)} \{match(u+i,v+j) + dist(u,i,v,j)\} \tag{2}$$

Where :

$$\beta(i) = \begin{cases} P_C - v & when\ i = 1 \\ 1 & otherwise \end{cases} \tag{3}$$

$$\begin{aligned} dist(u,i,v,j) = \ & f_l \cdot LD(u,i,v,j) \\ & - f_a \cdot AV(u,i,v,j) \\ & + f_{mp} \cdot MP(i,j) \end{aligned} \tag{4}$$

Eq.(3) shows that at the level of paragraph matching, only $1 \Leftrightarrow j$ and $i \Leftrightarrow 1$ matches are allowed. On the other hand, we do not place a specific limit on either $i$ or $j$, instead we allow 1 English paragraph to match with up to $P_C - v$ Chinese paragraphs (i.e., from $j$ th paragraph to the end of Chinese Corpus), and vice versa.

In Eq(2) and (4), $dist(u,i,v,j)$ is the distance between the Chinese texts from $u$ th to $(u+i-1)$ th paragraph and English texts from $v$ th to $(v+j-1)$ th paragraph. The larger the distance is, the less likely these two parts of texts are matched. And in Eq.(2) $match(u+i,v+j)$ is actually the sum of all the distances of the most possible matched pairs in the remaining paragraphs. Thus we choose $i$ and $j$ that can make the value of $match(u+i,v+j) + dist(u,i,v,j)$ to be smallest, and record it as the most possible match.

As shown in Eq.(4), $dist(u,i,v,j)$ is decided by three functions, and the factors $f_l$, $f_a$, $f_{mp}$ are constant numbers that can be determined through trial and errors. The first function $LD(u,i,v,j)$ is the length distance between this part of the English corpus and the Chinese corpus. As the previous works[5]-[10], we assume that each Chinese character is responsible for generating some number of English characters. After estimating some sample corpora, we get the expectation of this number $r = 2.61$, and the variance $\sigma^2 = 0.310$. Therefore we can compute $LD(u,i,v,j)$ by using a gaussian variable[6][7][9]:

$$LD(u,i,v,j) = \frac{l_e - l_c \cdot r}{\sqrt{l_c \sigma^2}} \tag{5}$$

Where $l_e$ and $l_c$ are respectively the length (measured by number of characters) of relative part of English and Chinese corpus.

The second function in Eq(4), $AV(u,i,v,j)$, is the sum of the value of matched anchors in both corpora. A negative sign is also attached before it as this function is to strengthen the match between the 2 strings. We define:

$$AV(u,i,v,j) = \sum_{MA} V(MA)$$

Where $MA$ means every pair of matched anchors in these two parts of English and Chinese texts.

Then comes the third function in Eq(4), $MP(i,j)$, which is a penalty for a match of multi paragraphs. We introduce it for a best balancing of the range of $dist(u,i,v,j)$ when $i$ , $j$ are varying. We know that the first function $LD(u,i,v,j)$ has a nice standard normal distribution. But the second function $AV(u,i,v,j)$ is monotonous, i.e. the more paragraphs involved, the larger $AV(u,i,v,j)$ would be. Sometimes when there are too many repeated anchors in consecutive paragraphs, it even improperly offsets the most hopeful decrease-and-increase variation of $LD(u,i,v,j)$. So we use the value of $AV(u,i,v,j)$ to define the penalty of multi-paragraph matching:

$$MP(i,j) = AV(u,i,v,j) \cdot (i + j - 2)$$

As a matter of fact, $MP(i,j)$ is just a fine tuning to the whole $dist(u,i,v,j)$, therefore its factor $f_{mp}$ should be much smaller than $f_a$.

## 3.4 Aligning Sentences and Clauses

After aligning the paragraphs, we can focus on how to align the sentences and clauses inside each pair of aligned paragraphs. For the sake of simplicity, we assume only 1 English paragraph matches with 1 Chinese paragraph from now on.

Because the sentence structure in English and Chinese are different, they are not so similar in the use of Sentence Mark. We regard clause as the basic unit of aligning in this step. Let $C_E$, $C_C$ represent the total number of clauses in the English paragraph and Chinese paragraph. Let $match(u,v)$ represent the total cost of matching from $u$ th clause to the end of the English paragraph with a string of Chinese text from the $v$ th clause to the end of Chinese paragraph, where $u = 1,2,\cdots,C_E$, $v = 1,2,\cdots,C_C$. In our program, we obtain $match(u,v)$ as follows:

$$match(u,v) = \min_{i=1}^{C_E - u} \min_{j=1}^{C_C - v} \{match(u+i, v+j) + dist(u,i,v,j)\} \quad (6)$$

Where:

$$\begin{aligned} dist(u,i,v,j) = \; & f_l \cdot LD(u,i,v,j) \\ & + f_{ms} \cdot MS(u,i,v,j) \\ & - f_a \cdot AVS(u,i,v,j) \\ & - f_{ws} \cdot WS(u,i,v,j) \end{aligned} \quad (7)$$

As we can see, there are some differences between Eq.(6)-(7) in this step and Eq(1)-(4) in last step. The first function $LD(u,i,v,j)$ in Eq(7) is similar to that in last step. The other functions which make differences are explained below.

To align clauses between an English paragraph and a Chinese paragraph, we allow an $m \Leftrightarrow n$ ($1 \le m < C_E - u$, $1 \le n < C_C - v$) match. Here we assume every part of Chinese has a translation in English. Actually, we found in some cases there are one or two Chinese clauses not translated, (As the corpora are free translation, we must clarify that when we say “not translated”, we mean the meaning of these Chinese clauses is not reflected in the English translation at all.) but what we are doing is to match a group of English clauses with a group of Chinese clauses, and this one or two not-translated Chinese clauses are usually among a can-be-matched group. If we allow these not-translated clauses to be extracted alone as a $0 \Leftrightarrow n$ match, we might divide and destroy the whole matched group. So we do not take the possibility of $0 \Leftrightarrow n$ match into consideration.

In Eq(7) when counting the distance, of course it is error-prone to allow matching of multi clauses in both languages. So we would like to give an advantage to the candidates of whole sentences. Therefore in Eq(7), we put a bonus of Whole Sentence $WS(u,i,v,j)$.

On the other hand, we must adopt some measures to prevent a match from involving too many clauses and sentences in either language, otherwise the final result would usually be only one match, i.e., the whole English paragraph matching with the whole Chinese paragraph, no detailed alignment at all. So we give a penalty to Multi-Sentence match, i.e. $MS(u,i,v,j)$.

$MS(u,i,v,j)$ will prevent a match containing inter-sentence clauses, i.e., prevent clauses in different sentences from appearing in the same matching group. In other words, we try to discourage matching with such a group of clauses that ends with a comma but has a period inside.

Of course we also make use of the anchors in this step. But the function regarding the anchors, $AVS(u,i,v,j)$, is much different with $AV(u,i,v,j)$ in last step. We define $AVS(u,i,v,j)$ as:

$$AVS(u,i,v,j) = S(u,i,v,j) \cdot \sum_{MA} V(MA) \quad (8)$$

This is because in the news-reports we can frequently find a sentence that lists a sequence of obvious anchors such as number data . Each clause in both languages would contain at least one of this sequence of anchors, and so we would like to match this kind of clauses as $1 \Leftrightarrow 1$, requiring it to give a penalty to other matches that contains too many clauses. But in other cases when there are not so many obvious anchors, the $1 \Leftrightarrow 1$ match is not so preferred, we would rather match a whole sentence. So we use $S(u,i,v,j)$ to measure the degree of Scattering of the anchors among these clauses.

$$S(u,i,v,j) = \left| \frac{S_E(u,i,v,j) - S_C(u,i,v,j)}{S_E(u,i,v,j) + S_C(u,i,v,j) + 0.1} \right| \qquad (9)$$

In Eq(9), $S(u,i,v,j)$ is determined by two terms, $S_E(u,i,v,j)$ and $S_C(u,i,v,j)$, which respectively represent the scattering of anchors in English and that in Chinese. We determine these two terms as follows:

$$S_E(u,i,v,j) = \sum_{MA} \left( ClNO_E(MA) - ClNO_E(MA_{PREV}) \right)$$
$$S_C(u,i,v,j) = \sum_{MA} \left( ClNO_C(MA) - ClNO_C(MA_{PREV}) \right)$$

Where $MA$ means every matched and not repeated pair of anchors in these two paragraphs. If an $MA$ is repeated in the current paragraph, it is undesirable to count it into the degree of scattering because we do not know which English anchor matches with which Chinese anchor in this case.

Each clause in a paragraph has a serial number (0,1,2...). $ClNO_E(MA)$ and $ClNO_C(MA)$ are respectively the Clause's numbers of this $MA$ in English and Chinese paragraph. $MA_{PREV}$ is the previous $MA$.

The scattering function $S(u,i,v,j)$ will be very important when all or most of the clauses contain matched anchors. If the order and distribution of the anchors in English are similar with that in Chinese, then $S_E(u,i,v,j)$ and $S_C(u,i,v,j)$ would be similar, thus $S(u,i,v,j)$ would be rather small, and so $AVS(u,i,v,j)$ would be small. Thus matching this entire pair sets will get less advantage compared to matching each clause one by one. On the other hand, if the order or the distribution of the anchors is too different, $S(u,i,v,j)$ will be quite large so that matching between this whole pair of sets gets more advantage.

For example, if we have 2 English clauses and 2 Chinese clauses, and each of them contains an anchor, as shown in Table 1.

| **Anchor** | **English** | **Chinese** |
|---|---|---|
| **Clause 1** | $A_{E,1}$ | $A_{C,1}$ |
| **Clause 2** | $A_{E,2}$ | $A_{C,2}$ |

**Table 1**

Suppose these clauses will be matched either as $1 \Leftrightarrow 1$ or as $2 \Leftrightarrow 2$. If the anchor $A_{E,1}$ matches with $A_{C,1}$, and $A_{E,2}$ matches with $A_{C,2}$ (we use $MA_1 = \{A_{E,1}, A_{C,1}\}$ and $MA_2 = \{A_{E,2}, A_{C,2}\}$ to represent these two pairs of matched anchors), let's see what will happen if we match the two English clauses with the two Chinese clauses in the entirety (i.e. $2 \Leftrightarrow 2$).

Knowing that $u = v = 1$, $i = j = 2$, we have:

$$\begin{aligned} S_E(1,2,1,2) &= ClNO_E(MA_2) - ClNO_E(MA_1) \\ &= 2 - 1 \\ &= 1 \end{aligned}$$

and

$$\begin{aligned} S_C(1,2,1,2) &= ClNO_C(MA_2) - ClNO_C(MA_1) \\ &= 2 - 1 \\ &= 1 \end{aligned}$$

So

$$\begin{aligned} S(1,2,1,2) &= \left| \frac{S_E(1,2,1,2) - S_C(1,2,1,2)}{S_E(1,2,1,2) + S_C(1,2,1,2) + 0.1} \right| \\ &= \left| \frac{1-1}{1+1+0.1} \right| \\ &= 0 \end{aligned}$$

Therefore $AVS(u,i,v,j) = 0$. That is to say, we cannot derive any advantage from these anchors if we match all these clauses together. Meanwhile the $1 \Leftrightarrow 1$ clause match will get a bonus from the matched anchors. Thus from the view of anchors, we will take the match of $1 \Leftrightarrow 1$ but not $2 \Leftrightarrow 2$ as the best match.

But on the other hand, if the anchors are cross-matching, i.e. $MA_1 = \{A_{E,1}, A_{C,2}\}$, $MA_2 = \{A_{E,2}, A_{C,1}\}$ the result would be much different.

$$\begin{aligned} S_E(1,2,1,2) &= ClNO_E(MA_2) - ClNO_E(MA_1) \\ &= 2 - 1 \\ &= 1 \end{aligned}$$

while

$$\begin{aligned} S_C(1,2,1,2) &= ClNO_C(MA_2) - ClNO_C(MA_1) \\ &= 1 - 2 \\ &= -1 \end{aligned}$$

Thus

$$S(1,2,1,2) = \left| \frac{S_E(1,2,1,2) - S_C(1,2,1,2)}{S_E(1,2,1,2) + S_C(1,2,1,2) + 0.1} \right|$$

$$= \left| \frac{1 - (-1)}{1 + (-1) + 0.1} \right|$$

$$= 20$$

Therefore the $2 \Leftrightarrow 2$ match gets a much larger bonus than the $1 \Leftrightarrow 1$ match. So we will take the match of $2 \Leftrightarrow 2$ as the better matching result.

This is the simplest example. In the more complicated cases, function $S(u,i,v,j)$ also acts as a good role to judge the scattering of the matched anchors.

Fig 1-3 show three typical kinds of CNS news texts aligned by our program.

In Fig1, there are no obvious anchors, so we can only use the length of clauses as a clue to align them. As we can see, a period in English is used to break a complete Chinese sentence into two parts.

In Fig2, in the first pair of sentences, there are some obvious anchors. But these anchors are cross-matching, so we match the complete sentence as a whole.

In Fig3, there are quite a lot of obvious anchors, and all of these anchors scatter evenly in the same order in both texts. So we divide the sentences into clause-to-clause match.

## 4. Experimental Result

Altogether, we have tested this algorithm with 20 pairs of randomly selected news files. The basic data of these files are shown in **Table 2.**

| | English | Chinese |
|---|---|---|
| Files | 20 | 20 |
| Paragraphs | 126 | 153 |
| Sentences | 297 | 314 |
| Clauses | 782 | 844 |
| Characters | 35,284 | 16,106 |

**Table 2** Data of news files

The accuracy of the algorithm is measured by dividing the number of correct pairs by the total pairs obtained from the algorithm . We have subjected these files to three different testing methods. The first was done by only allowing length measures to govern the alignment process. The second was done by only allowing anchors, and in the last method the whole function was used, namely, both length and anchors. The results are tabulated in **Table 3**.

Of the 20 news files, the number of paragraphs aligned correctly by each of the three methods were 77, 72, 103, respectively, as shown in Row(1) of **Table 3**. While the total number of paragraph pairs were similar by the three methods as shown in Row(2), it does not necessarily mean the matches were similar. Actually the first two methods made most of the mistakes at different places. Since we allowed match of multi-paragraph, it is believed that the similarity of the total number of paragraph pairs is occasional. This indicates that both length measures and anchors are needed to produce good alignment results on the level of paragraph. Examination of the contents shows that where translation patterns in the bilingual texts are quite regular (such as without abruptly omitted Chinese paragraphs), length measure alone would be sufficient. On the other hand, when there are plenty of anchors, the anchors can become a dominating factor.

Next in Row(4)-(6) it is seen that not only the number of correct pairs, but also that of total pairs produced are quite different among the three methods. Using length measure alone lacks the ability to divide long sentences into short clause-to-clause alignment, so the total pairs produced is much less than the other two methods, as shown in Row(5). But when we say a pair is correct, we mean the contents of this pair in both language are best matched, without any part of one language actually matching with a part of another pair in the other language. So we might call a pair of sentences a correct pair even if it can be divided into many clause-to-clause pairs. As we can see in Row(6), the only acceptable accuracy is by using both length and anchor measures.

Noting that quite a number of wrong clause pairs exist in those wrong paragraph pairs (actually all the clause pairs produced from the wrong paragraph pairs should be considered wrong), we also count the clause pairs produced from the correct paragraph pairs only, listed in Row(7)-(9). It is interesting to see in Row(9) that given the correct alignment of paragraphs, the accuracy of using length measure alone and using anchor measure alone both increases a lot. And the accuracy of using both measures in this case hits a high of 96.41%. Considering the texts are free translation, this is quite a satisfactory result.

## 5. Conclusion

We have presented an algorithm for automatic alignment of English and Chinese bilingual CNS news. While works on automatic alignment on other text pairs have been reported in the literature, our work here is unique in several respects. Firstly, the method handles two texts that are widely different in sentential structure. Secondly, because of the sentential structure differences, we have decided not to constrain matching at sentence boundaries as in the earlier works. Thirdly, we avail ourselves of the feature of news reports, namely occurrence of common anchors, and make it a lexical cue to be combined with statistical approach to get a best alignment. Finally, by measuring the scattering of the anchors, we can achieve a clause-to-clause match in some parts of the texts.

Because we allow matching with clauses, matching was more difficult and error-prone. The use of both length measures and anchor measures has, however, proven to be a remarkable means of guiding the automatic alignment with an accuracy rate up to more than 90% in corpora of free translation.

Up to now we only use the numbers and dates and places as the anchors. We believe that if we adopt some other cues that are relatively fixed translation (one cue is words such as "银银" and "bank", another cue is specific characters that will not changed after translation), we can improve the accuracy even more.

We have mentioned that the texts we are using are taken from CNS webpage. However on their webpage the files in the two languages are too small, and in addition, are not placed in the same order. Therefore it is somehow tedious to match the files according the titles on their web page and download them by hand. As a future work, we are thinking about automatically download these files, and then use a similar algorithm as described in this paper to match them. Of course this matching work is much simpler because we only need to deal with $1 \Leftrightarrow 1$ match.

# 6. Acknowledgements

Special thanks are due to China News Service for their permission to use their news reports.

| | | Using Length Measure | Using Anchors | Using both Length and Anchor |
|---|---|---|---|---|
| 1 | No. of correct paragraph pairs | 77 | 72 | 103 |
| 2 | No. of total paragraph pairs | 103 | 107 | 106 |
| 3 | Accuracy rate of paragraph matching | 74.8% | 67.3% | 97.2% |
| 4 | No. of correct clause pairs | 127 | 186 | 242 |
| 5 | No. of total clause pairs | 213 | 270 | 261 |
| 6 | Accuracy rate of clause matching | 59.6% | 68.9% | 92.7% |
| 7 | No. of correct clause pairs in correct paragraph pairs | 127 | 165 | 232 |
| 8 | No. of total clause pairs in correct paragraph pairs | 152 | 203 | 251 |
| 9 | Accuracy rate of clause matching in correct paragraph pairs | 83.6% | 81.2% | 96.41% |

**Table 3** Aligning results obtained from 3 different testing methods.

| | |
|---|---|
| The vice president noted that judicial review corrected and contained the improper application of administrative power while at the same time eliminating or reducing negative impacts on the state administration. | 这这这这这这这这这这这这这这这这这<br>判判判判判判判判判判判判判判判判判<br>当当当当当当当当当当当当当当当当当<br>家家家家家家家家家家家家家 |
| It also protects and supports the intra vires acts of the administration. | 也也也也也也也也也也也也也也也也也<br>职职职 |

**Fig.1**

| | |
|---|---|
| Mr. Luo said that 141,949 cases were handled by the Administrative Appeals Tribunals, between October 1990 and June 1995. | 罗罗罗罗罗罗罗罗罗罗罗罗罗罗罗罗罗<br>月月月月月月月月月月月月月月月月月<br>行行行行行行行行行行行行行行行行 |
| Decisions appealed came from 40 administrative areas including land, public security, urban construction, industry and commerce, environmental protection, prices, finance, customs, forestry, mining, taxation and technological supervision. | 案案案案案案案案案案案案案案案案案<br>、、、、、、、、、、、、、、、、、<br>产产产产产产产产产产产产产产产产产<br>理理理理 |

**Fig.2**

| | |
|---|---|
| In the first nine months of this year, | 今今今今今今今今 |
| China's foreign trade reached US$ 197.66 billion, | 中中中中中中中中中中中中中中中中中<br>六六六六六六六六 |
| or 25.5 per cent up when compared with the same period last year. | 比比比比比比比比比比比比比比比比 |
| Exports were valued at US$ 107.04 billion, | 其其其其其其其其其其其其其其其其 |
| an increase of 34.8 per cent, | 增增增增增增增增增增增 |
| and imports at US$ 90.62 billion, | 进进进进进进进进进进进进进 |
| an increase of 16.2 per cent. | 增增增增增增增增增增 |
| The trade surplus stood at US$ 16.42 billion. | 外外外外外外外外外外外外外外外外 |

**Fig.3**